\documentclass[twocolumn]{elsarticle}

\usepackage[margin=1.8cm]{geometry}
\usepackage{multicol}

\usepackage{amsmath}
\usepackage{amsfonts}
\usepackage{amssymb}
\usepackage{natbib}
\usepackage{url}
\usepackage{hyperref}
\usepackage[linesnumbered,ruled,vlined]{algorithm2e}

\usepackage[dvipsnames]{xcolor}
\usepackage[normalem]{ulem}

\journal{Energy and Buildings}

\title{Occupancy Detection Based on Electricity Consumption}

\author[hw]{Thomas Brilland}

\author[hw]{Guillaume Matheron\corref{mycorrespondingauthor}}
\cortext[mycorrespondingauthor]{Corresponding author}
\ead[url]{www.hellowatt.fr}
\ead{guillaume_pub@matheron.eu}

\author[hw]{Laetitia Leduc}

\author[hw]{Yukihidé Nakada}

\address[hw]{Hello Watt, 48 rue René Clair 75018 Paris, France}

\begin{document}
    \begin{abstract}
        This article presents a new methodology for extracting intervals when a home is vacant from low-frequency electricity consumption data.
        The approach combines multiple algorithms, including change point detection, classification, period detection, and periodic spikes retrieval.
        It shows encouraging results on both simulated and real consumption curves.
        This approach offers practical insights for optimizing energy use and holds potential benefits for residential consumers and utility companies in terms of energy cost reduction and sustainability.
        Further research is needed to enhance its applicability in diverse settings and with larger datasets.
    \end{abstract}
    \twocolumn[
        \begin{@twocolumnfalse}
            \maketitle
        \end{@twocolumnfalse}
    ]


    \section{Introduction}

    \paragraph{Energy consumption}
    In the wake of the 2021-2023 energy crisis that started in Europe, controlling electricity consumption has become a major challenge for European and French consumers who saw prices increase drastically in 2022 \citep*{noauthor_energy_2023,noauthor_electricity_nodate}.
    Knowing the sources of electricity consumption in residential homes, and how to reduce them, is an effective way to reduce bills and ecological footprints.
    Smart meters, which have been massively developed in many countries over the past few years, can provide a more detailed understanding of individual consumption, enabling optimization \citep*{noauthor_cre_nodate}.
    In France, more than 34 million of these meters, known as Linky, were installed by the end of 2021 \citep*{ademe_compteur_2021}.
    These smart meters record total electricity consumption data every 30 minutes and send it to the Enedis infrastructure \citep*{horowitz_power_2008}.
    This low sampling rate makes signal disaggregation a difficult task \citep*{culiere_bayesian_2020, belikov_domain_2022}.

    \paragraph{Occupancy detection}
    Identifying holiday intervals, which we define as intervals of several days when inhabitants are away, can be useful for disaggregating the consumption of multiple devices.
    These intervals can help disaggregate the consumption of devices that are typically left active during the holidays such as fridges and water heaters.
    Moreover, identifying when a home is unoccupied can improve existing disaggregation models by filtering irrelevant consumption data from the training data set of supervised models, or be used as a prior filtering step before unsupervised models are applied.
    The more precise the disaggregation, the more control consumers have over their consumption.
    Access to a labelled dataset with information about holidays is tedious to obtain on a large scale, making supervised learning difficult \citep*{kleiminger_eco_2016}.

    \section{Background}
    In this section we describe the concepts and notations used in this article.

    \paragraph{Time series}
    A \emph{time series} is a discrete sequence of real values indexed by time.
    In this article we assume that these points are equally spaced with distance $T_s$, called the \emph{sampling period}.
    Its reciprocal $f_s = \frac{1}{T_s}$ is called the \emph{sampling frequency} or \emph{sampling rate}.

    \newcommand{\signal}[3]{\{#1\}_{#2}^{#3}}
    \newcommand{\series}[3]{#1_{#2..#3}}
    Let $y = \signal{y_t}{t=0}{T}$ be a time series defined between 0 and $T-1$.
    In the remainder of this article, it will be denoted $\series{y}{0}{T}$.
    More generally, $\series{y}{a}{b}$, with $0 \leq a < b \le T$, will refer to the subsequence $\signal{y_t}{t=a}{b}$ consisting of $y_t$ for $a \le t < b$. 

    \paragraph{Periodicity, mean, and variance}
    A time series $y$ is said to be \textit{periodic} of period $P$ if $y(t) = y(t+P)$ for all $t$
    or, equivalently, if $\series{y}{a}{b} = \series{y}{a+P}{b+P}$ for all $0 \le a < b \le T-P$.

    Given a time series $\series{y}{a}{b}$, the mean value of the time series between $a$ and $b-1$ will be denoted by $\overline{\series{y}{a}{b}}$ and its variance will be denoted by $\sigma_{\series{y}{a}{b}}^{2}$, 

    \paragraph{Discrete Fourier Transform}
    The Discrete Fourier Transform (DFT) is a fundamental tool that allows for the decomposition of a finite time series into a sum of periodic components at different frequencies.
    For a time series $\series{y}{0}{T}$, it is defined as a $T$-long sequence of complex numbers $Y(f)$ as follows:
    \begin{equation}
        Y(f_{k/T}) = \sum_{t=0}^{T-1} y_t e^{- \frac{2\pi i kt}{T}}, \quad k = 0, 1, 2, \ldots, T-1
    \end{equation}
    where $f_{k/T} = \frac{2\pi k}{T}$ corresponds to the frequency captured by each component.

    We can obtain the contribution of each frequency to the time series with the periodogram
    \begin{equation}
        P(f_{k/T}) = \left\lVert Y(f_{k/T})\right\rVert ^{2}.
    \end{equation}
    This is a way to find dominant periodicities in a time series \citep*{vlachos_periodicity_2005}.

    \paragraph{Autocorrelation}
    Autocorrelation is a measure of similarity between a time series and a lagged version of itself, and is another tool to find dominant periodicites in a time series. 
    The autocorrelation value at lag $k$ for a time series $\series{y}{0}{T}$ is computed as follows:
    \begin{equation}
        ACF(k) = \frac{1}{T} \sum\limits_{t=0}^{T-1} y_t y_{t+k}.
    \end{equation}

    \paragraph{Spike Detection}
    A \emph{spike} is defined as a contiguous subsequence of a time series where all values are above a given threshold.
    The task of spike detection includes automatic threshold selection and enumeration of the spikes rising above this threshold.
    Different features can be defined for each spike: start time, end time, and maximum rise above threshold, for example.

    \paragraph{Gaussian distribution}
    The Gaussian distribution is characterized by its bell-shaped curve, and is defined by the probability density function
    \begin{equation}
        f(x) = \frac{1}{\sqrt{2\pi}\sigma}e^{-\frac{(x-\mu)^{2}}{2\sigma^{2}}}
    \end{equation}
    with $\mu$ the mean of the distribution and $\sigma^{2}$ its variance.

    \paragraph{Occupancy / Holidays}
    In this article, we define \emph{occupancy} as the times when a home is occupied by at least one person, and a \emph{holiday} to be an interval of at least three days when a home is unoccupied. 
    Holidays detection thus consists in finding long intervals (at least three days) when inhabitants left their home empty.

    \paragraph{Classification Metrics}
    We will use fundamental classification metrics to evaluate model performance.

    \textbf{Precision:} Precision is the ratio of true positive predictions to the total number of positive predictions made by a classification model.
    It measures the accuracy of positive predictions.
    \begin{equation}
        \text{Precision} = \frac{\text{True Positives}}{\text{True Positives + False Positives}}
    \end{equation}

    \textbf{Recall:} Recall is the ratio of true positive predictions to the total number of actual positive instances in the dataset.
    It measures the ability of the model to identify all positive instances.

    \begin{equation}
        \text{Recall} = \frac{\text{True Positives}}{\text{True Positives + False Negatives}}
    \end{equation}

    \textbf{F1-score:} The F1-score is the harmonic mean of Precision and Recall.
    It provides a balance between those two metrics, especially in our case for which there is an imbalance between the number of positive instances and the number of negative ones.

    \begin{equation}
        \text{F1-score} = 2 \times \left(\frac{\text{Precision} \times \text{Recall}}{\text{Precision + Recall}}\right)
    \end{equation}

    \section{Related Work}

    \paragraph{Signal Disaggregation}
    Nonintrusive Load Monitoring (NILM) was introduced by G.W. Heart in 1992 \citep*{hart_nonintrusive_1992}, and is an active research area.
    Disaggregation of energy consumption has thus been extensively explored, with various possible approaches that can obtain solid results \citep*{kaselimi_towards_2022}.
    Most studies use supervised learning methods, with high-frequency disaggregated power load available with annotations \citep*{kolter_redd_nodate}.
    For example, the authors of \citep*{zhou_non-intrusive_2021} use the UK-DALE dataset \citep*{kelly_uk-dale_2015} to train a CNN-LSTM deep learning model.

    \paragraph{Occupancy Detection}

    There are many different methods for home occupancy detection, targeted to both residential homes and commercial buildings.
    Multiple sensors have been used, such as cameras \citep*{callemein_anyone_2019} and infrared \citep*{singh_non-intrusive_2019}, but also CO2 sensors \citep*{zhou_novel_2020} and even Wi-Fi signals \citep*{wang_inferring_2019}.
    These studies apply machine learning and deep learning models to the data collected by their sensors.
    The authors of \citep*{wang_inferring_2019} apply various models (Neural Network, Random Forest, and LSTM) to Wi-Fi data to find the occupant counts of a building.

    Some methods also use electricity consumption, and NILM methods, sometimes combined with water consumption data in order to detect occupancy \citep*{vafeiadis_machine_2017,chen_non-intrusive_2013, kleiminger_occupancy_2013, kleiminger_household_2015}.
    In \citep*{vafeiadis_machine_2017}, the researchers collected their own data on water consumption and electric consumption during a month, and trained multiple machine learning models for occupancy detection.

    However, our use case is distinct because we are restricted to low-frequency electricity consumption data.
    Our target is an unlabelled dataset with diverse contexts such as different types of houses and different numbers of inhabitants.
    In addition, we are looking for relatively long vacancy periods instead of short vacancies within a day, or occupant counts.
    Finally, our algorithm aims at extracting specific consumption for devices that stay activated during unoccupied periods.

    \section{Methodology}

    \begin{figure*}
        \centering
        \includegraphics[scale=0.6]{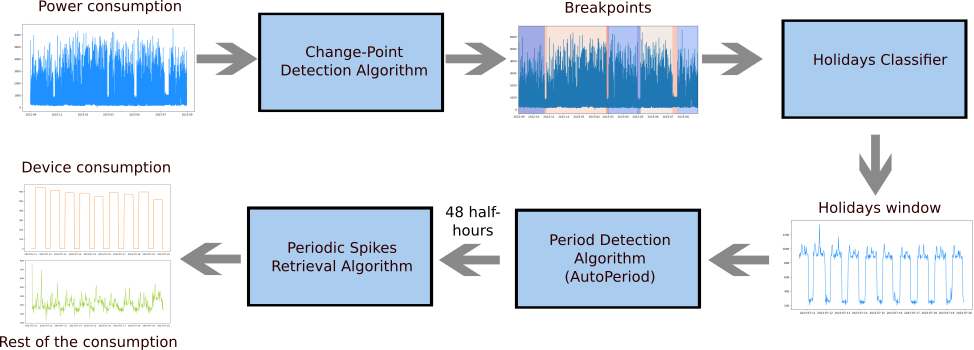}
        \caption{The algorithm for holiday detection}
        \label{fig:cpd}
    \end{figure*}

    In this article, we combine different methods to detect potential holiday intervals, and then disaggregate device signals during these intervals using periodicity detection.
    Once a period has been detected, the spike detection algorithm is applied to the signal to retrieve the periodic spikes pattern in the aggregated signal.

    \subsection{Problem Formulation}

    The input of the problem is the electricity consumption for a home, a one-year long time series $\series{y}{0}{T}$ with a sampling period of 30 minutes.

    The objective is to extract a set of pairs of points that correspond to extended holiday periods, that is, a set
    $ \{(a_0, b_0), (a_1, b_1), \ldots, (a_n, b_n)\} $ such that for every $k$ in $[0, n]$, $\series{y}{a_k}{b_k}$ represents a holiday period.

    \subsection{Change Point Detection}
    The first part of our algorithm for holiday detection consists in separating the input signal into several segments in order to identify potential holiday candidates.
    This corresponds to the Change Point Detection problem, which is an extensively researched area in signal processing~\citep*{truong_selective_2020, aminikhanghahi_survey_2017}.
    For our case, where we need an efficient offline algorithm that works for unlabelled data, the Bottom-Up algorithm is a good candidate. 

    This algorithm requires a cost function that measures time series homogeneity.
    The idea is to split the input signal into small intervals and iteratively merge consecutive intervals if their union is sufficiently homogeneous, which we measure by comparing the homogeneity of the two intervals separately with their union via the cost function.
    We do this until a stopping criteria is met. 
    The algorithm (Algorithm \ref{algo:botup}) requires the starting intervals size (grid size) as input, and a stopping criterion that can be the number of windows or a threshold for the maximum cost of a merger.

    \begin{algorithm}[!ht]
        \KwIn{Signal $\series{y}{0}{T}$, cost function $c(\cdot)$, stop criterion, grid size $\delta$}

        $L \gets \{t_0, t_1, \ldots, t_n\} $ with $t_i = i\delta, t_n = T$ \\
        \While{stopping criterion not met}
        {
            $k \gets |L|$ \\
            $\widehat{i} \gets $argmin$_i c(\series{y}{t_{i-1}}{t_{i+1}}) - [c(\series{y}{t_{i-1}}{t_{i}}) + c(\series{y}{t_{i}}{t_{i+1}})]$ \\
            Remove $t_{\widehat{i}}$ from $L$
        }

        \KwOut{$L$: List of indexes for the found breakpoints}

        \caption{Bottom-Up Algorithm}
        \label{algo:botup}
    \end{algorithm}

    \begin{figure}
        \centering
        \includegraphics[scale=0.18]{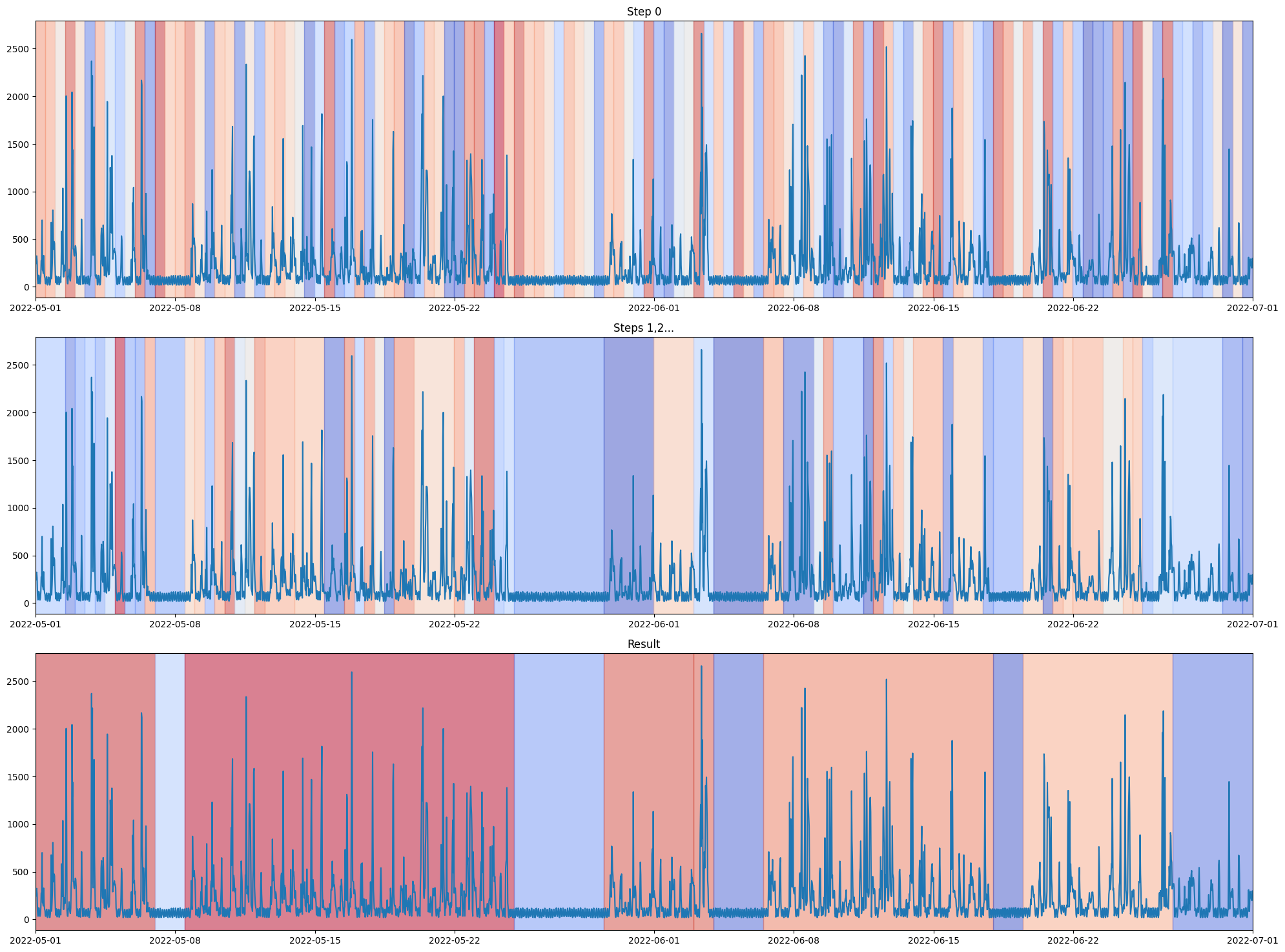}
        \caption{\centering Bottom-Up Algorithm step by step. Colors help visualize the breaking points.}
        \label{fig:cpd}
    \end{figure}

    There are multiple existing cost functions for this method.
    We chose the cost function that approximates our signal to a Gaussian with piecewise constant mean and variance, allowing us to detect changes in the sliding mean and the sliding variance.

    To detect changes in the rolling average of a signal, the appropriate cost function is $c_{L_2}$, defined as
    \begin{equation}
        c_{L_2}(\series{y}{a}{b}) = \sum\limits_{t=a}^{b-1}\left\lVert y_t - \overline{\series{y}{a}{b}}\right\rVert_{2}^{2}.
    \end{equation}

    In order to detect changes in variance as well, the cost function becomes
    \begin{equation}
        c_\Sigma(\series{y}{a}{b}) = (b - a) \log(\sigma^{2}_{\series{y}{a}{b}}) + \frac{1}{\sigma^{2}_{\series{y}{a}{b}}} \sum\limits_{t=a}^{b-1}\left\lVert y_t - \overline{\series{y}{a}{b}}\right\rVert_{2}^{2}.
        \label{eq:cost}
    \end{equation}

    \subsection{Holiday Classification}
    Once the signal is split into segments (or intervals), each one needs to be classified as holiday or occupied.
    In order to do this, different classifiers have been tested.

    First, it needs to be noted that we can't just alternate between occupied and vacant, because there can be change points between two occupied segments.
    However there are generally no change points between two consecutive holiday segments.
    We consider two versions of the algorithm: compare the studied segment to the whole signal, or only to its neighboring segments.
    There are also two criteria $\gamma$ to compare: mean consumption or the variance of the consumption.
    A segment is classified as holiday if its mean or variance is significantly lower than what it is being compared to.

    When it is compared to its neighbors, a segment is classified as a holiday if the criteria $\gamma$ on the segment is inferior to $\gamma$ on both of its neighboring segments (\ref{eq:constraint1}) and to the average value multiplied by a certain ratio $r$ (\ref{eq:constraint2}):
    \begin{align}
        & \gamma({\series{y}{t_k}{t_{k+1}}}) \le \, \textrm{min}(\gamma(\series{y}{t_{k-1}}{t_k}), \gamma(\series{y}{t_{k+1}}{t_{k+2}})) \ ; \label{eq:constraint1} \\
        & \gamma({\series{y}{t_k}{t_{k+1}}}) \le \, \frac{r}{2} (\gamma(\series{y}{t_{k-1}}{t_k})+\gamma(\series{y}{t_{k+1}}{t_{k+2}})) \ . \label{eq:constraint2}
    \end{align}
    When it is compared to the whole signal, it is classified as holiday if
    $\gamma({\series{y}{t_k}{t_{k+1}}}) \leq \gamma({\series{y}{0}{T}})$.

\begin{figure}
        \centering
        \includegraphics[scale=0.33]{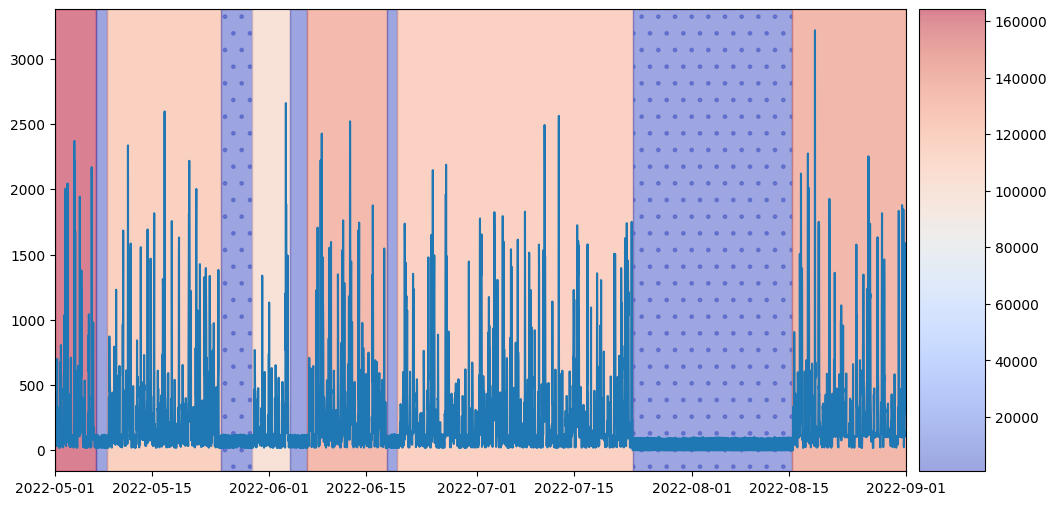}
        \caption{\centering Holiday Classifier $F_{var}$ applied to a 2-year signal. Only the last 4 months are shown here. Windows classified as holiday have dots. Background color corresponds to the window variance.}
        \label{fig:holiday}
    \end{figure}

    Table \ref{table:holiday_classifiers} summarizes the four classifier versions that are tested in this article.
    Figure \ref{fig:holiday} shows an example of results using $F_{var}$ classifier.

    \begin{table}
        \centering
        \begin{tabular}{|c|c|c|}
            \hline
            Version Name & Comparison & Criteria \\
            \hline
            $N_{mean}$   & Neighbours & mean     \\
            \hline
            $N_{var}$    & Neighbours & variance \\
            \hline
            $F_{mean}$   & Full       & mean     \\
            \hline
            $F_{var}$    & Full       & variance \\
            \hline
        \end{tabular}
        \caption{Holiday Classifier Versions}
        \label{table:holiday_classifiers}
    \end{table}

    \subsection{Period Detection}
    In order to detect periodicity in segments identified as holidays, we applied the AutoPeriod method described in \citep*{vlachos_periodicity_2005}.

    First, we apply a Discrete Fourier Transform to our signal, and identify the frequencies corresponding to significant spikes in the periodogram. 
    To determine significance, we compute 100 different random permutations of the input signal, compute the Fourier transform and its periodogram for each permutation, and set the threshold as the second-highest value obtained across all permutations. 
    The underlying idea is that periodicities in the original signal should be significantly more prominent than periodicities in any permutation. 
    
    Next we proceed to compute the autocorrelation of the signal.
    For each candidate frequency, we verify whether its corresponding period aligns with a local maximum in the autocorrelation, employing a method involving two linear regressions.
    If it does, and if this local maximum is significant, the candidate period is considered valid.
	
\begin{figure}
        \centering
        \includegraphics[scale=0.2]{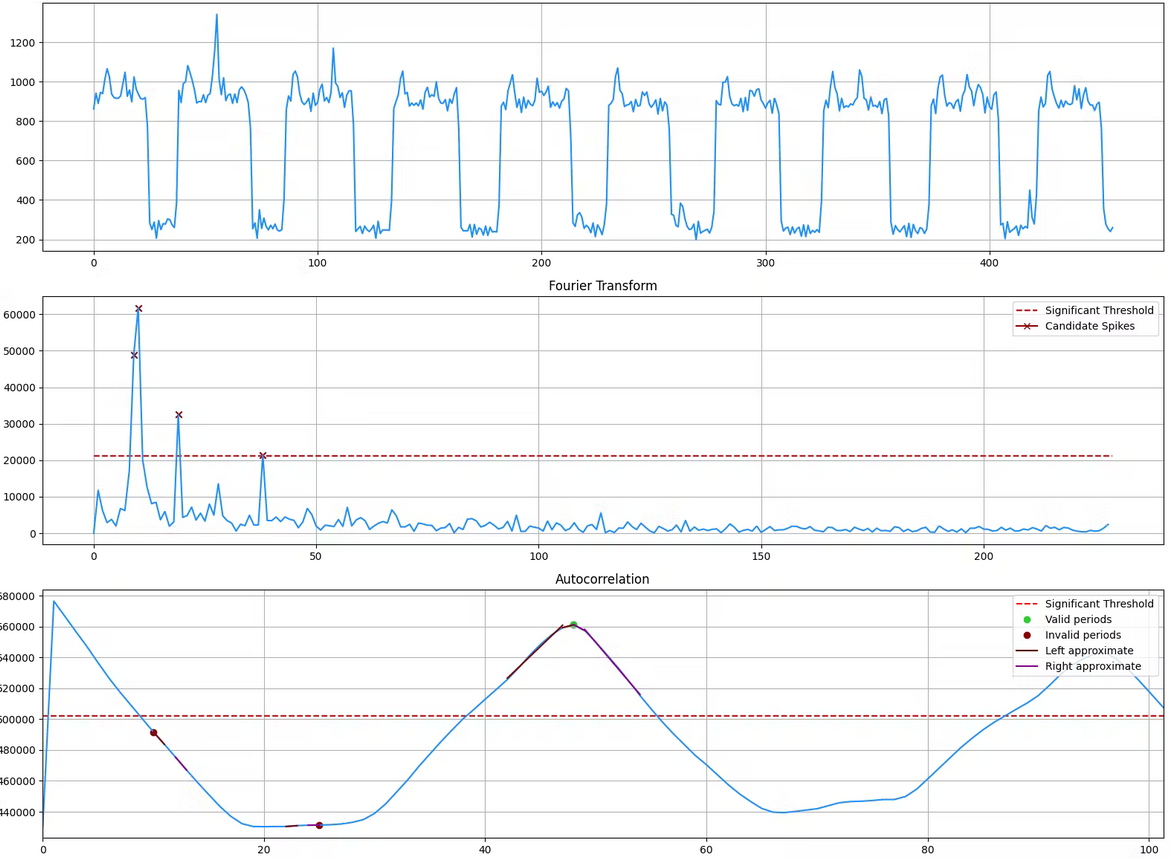}
        \caption{\centering Period Detection method for a noisy asymmetric squared signal. The second plot corresponds to the DFT, and the third one is Autocorrelation. Significance thresholds are in red. The only valid period is 48, the others are discarded.}
        \label{fig:autoperiod}
    \end{figure}

	Figure \ref{fig:autoperiod} illustrates the period detection method for a noisy asymmetric squared signal.

    \subsection{Periodic Spikes Retrieval}
    Once periodicity in a signal that corresponds to an unoccupied home was detected, we want to extract its periodic parts, which correspond to one or more devices that were left activated.

    First, the spikes in our signal need to be detected.
    We apply a Kernel Density Estimator to our signal, with a Gaussian kernel, in order to compute the threshold above which points are considered as spikes.
    We use the same algorithm described in \citep*{belikov_domain_2022}.

    Once the list of spikes has been retrieved, the differences between the start indexes for each spike are computed, referenced as $\Delta s_k$ in Algorithm \ref{algo:retrieval}.
    The maximum alignment error $m$ denotes the maximum offset allowed from perfect periodicity.
    To retrieve the periodic pattern, the best starting spike $s_i$ is determined.
    Then the algorithm scans the entire signal backwards and forwards, by making jumps of distance $P$ or $-P$ (the detected period) and retrieves the closest spike, if the error is lower than $m$.

    \begin{algorithm}[!ht]
        \KwIn{Start time of spikes ${\{s_k\}}_0^n$, Period $P$, max error $m$}

        $\Delta s_k \gets {\{ s_k - s_i\}}_{i=0}^{n}$ \\
        $D \gets \max \{ \max |\Delta s_k| \}_{k=0}^{n} $ 
        
        $i \gets \operatorname{argmax}\{\lvert \{\Delta s_k \mod T \leq m\} \rvert\}_{k=0}^{n}$ \\
        $S \gets \{s_i\}$ \\
        \For{$t \in \{-P, P\}$}
        {
            $k \gets i$\\
            $j \gets 1$ \\
            \While{$j \le \lfloor \frac{D}{P} \rfloor $}
            {
                \If{$\min \{\lvert \Delta s_k - jt \rvert\} \leq m$}
                {
                    $k \gets \operatorname{argmin}\{\lvert \Delta s_k - jt \rvert \}$ \\
                Add $s_k$ to $S$ \\
                    $j \gets 1$
                }
                \Else{
                    $j \gets j+1$
                }
            }
        }

        \KwOut{List of periodic spikes $S$}

        \caption{Periodic Spikes Retrieval}
        \label{algo:retrieval}
    \end{algorithm}

    After the list of spikes has been retrieved, the periodic signal is built with each spike's average value.
    This value corresponds to the average value of the signal during the spike minus the baseline value, which is the mean value of the points just before and just after the spike.

    \section{Results}

    \subsection{Dataset Presentation}
    The dataset used in this paper comes from the data collected by Hello Watt.
    Users authorize Hello Watt to collect their home power usage at a resolution of 30 minutes.
    Our dataset is a random sample of 140 homes among Hello Watt users for which more than 90\% of the data between September 2021 and September 2023 is available. 
    This dataset has been used to evaluate our results for the full algorithm.

    For period detection and signal retrieval, a set of synthetic signals for various cases has also been generated.
    They consist of asymmetric squared signals with gaussian noise and random spikes added.

    \begin{figure}
        \centering
        \includegraphics[scale=0.33]{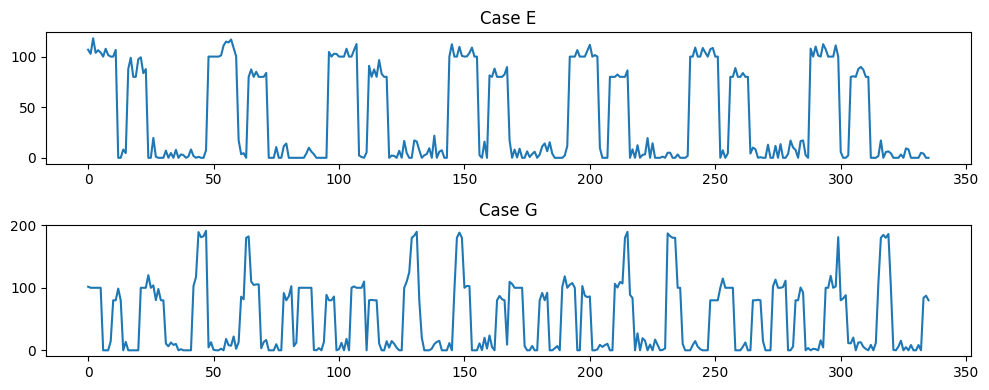}
        \caption{\centering Examples E and G of the synthetic signals that were generated.}
        \label{fig:signal}
    \end{figure}

    \begin{table}
        \centering
        \resizebox{0.45\textwidth}{!}{%
            \begin{tabular}{|c|c|c|c|c|}
                \hline
                Case Name & Period 1 & Period 2 & Random Spikes \\
                \hline
                A         & 48       & None     & No          \\
                \hline
                B         & 48       & None     & Yes           \\
                \hline
                C         & 23.6     & None     & No            \\
                \hline
                D         & 23.6     & None     & Yes           \\
                \hline
                E         & 48       & 48       & No            \\
                \hline
                F         & 48       & 48       & Yes           \\
                \hline
                G         & 21       & 17       & No            \\
                \hline
                H         & 23.6     & 29.4     & No            \\
                \hline
            \end{tabular}%
        }
        \caption{\centering Generated synthetic signals. Period 1 and Period 2 describe the period of the periodic components. If Random Spikes is Yes, a random component is added.}
        \label{table:signal}
    \end{table}

    In Figure \ref{fig:signal} are displayed two of the eight cases described in Table \ref{table:signal}.
    When Period 2 is not None, the second signal is set to have a lower maximum value and shorter spikes.
    Its starting point is offset by 10.
    When Random Spikes is not None, the probability of appearance of a spike is set with a Poisson distribution with a probability of 0.06, and their value are equal on average to half of the first signal.

    In the first 4 cases (A to D), there is only one asymmetric square signal, with gaussian noise and additional random spikes for cases B and D.
    In cases A and B the periodicity is set to the sampling rate, while in C and D it is not.
    In the last 4 cases, there are two signals.
    In cases E and F the signals have the same period, proportional to the sampling period.
    In case G, the two signals have different periods, proportional to the sampling period.
    In case H, the periods are different and both are offset from the sampling period.

    \subsection{Change Point Detection}
    The authors of \citep*{truong_selective_2020} implemented the Python library \texttt{ruptures} that was used for this paper.
    The stopping criterion for the Bottom-Up Algorithm is the maximum value of the cost function below which the algorithm keeps merging consecutive windows.
    To select this value, two criteria were used: the minimum window size for a signal has to be around 3 days (our definition of holidays), and the number of windows has to be on average similar to the number of windows if the home is vacant during school holiday in France and occupied otherwise.
    Then the Bottom-Up algorithm was tested with various maximum thresholds for 40 random households: the best threshold to fit our criteria on average was 250. \\

    \subsection{Data Labeling}
    In order to evaluate the change point and the holiday detection algorithms, the results of the Bottom-Up algorithm on the dataset have been labeled.
    Each window has been manually labeled as vacant or occupied based on relative consumption pattern during the two years.
    Out of the 140 homes, 65 of them had ``perfect'' results: all the vacant windows were captured by the change-point detection.
    29 of them had no visible vacant sub-segment, thus the change-point detection algorithm could not have detected any window.
    22 of them had some missing windows, meaning they had at least one visible vacant window that was not captured properly.
    For these windows, the holiday classifier could not classify them correctly since it has not be retrieved by the change-point detection algorithm.
    Finally, 24 cases were too difficult to label in a non-arbitrary way, so we decided to put them aside.

    \subsection{Holiday Classification}
    Since the classifiers are not evaluated on the 24 particular cases, it leaves 116 households in the dataset.
    The holiday classifiers need a ratio parameter as input.
    We used cross-validation process to fit the ratio parameter on a part of the dataset (20\%) and measure their performance on another part (80\%).
    The best ratio is chosen based on the F1-score on the fit part of the dataset.
    In order to compare our classifiers to a baseline, we used a binomial classifier, that classifies each window as holidays with a probability of 0.25 (the proportion of holiday labels in the dataset). \\
    Table \ref{table:results} presents the average results on 5-fold cross-validation.
    Each of our classifiers has significantly better results than a random classifier.
    Variance-based classifiers perform better than Mean-based ones, and the best performing classifier is the $F_{var}$ version.
    Comparing the variance of the window to the full signal seems to be the best method to classify holidays.

    \begin{table}
        \centering
        \begin{tabular}{|c|c|c|c|}
            \hline
            Version    & Precision      & Recall          & F1-score \\
            \hline
            Random     & 0.165          & 0.131           & 0.146     \\    
            \hline
            $N_{mean}$ & 0.76           & 0.849           & 0.798     \\
            \hline
            $N_{var}$  & 0.842          & \textbf{0.919}  & 0.877     \\
            \hline
            $F_{mean}$ & 0.804          & 0.768           & 0.771     \\
            \hline
            $F_{var}$  & \textbf{0.904} & 0.907           & \textbf{0.904}      \\
            \hline

        \end{tabular}
        \caption{Holiday Classifiers Performance}
        \label{table:results}
    \end{table}

    \subsection{Period Detection and Spikes Retrieval}

    We applied the Period Detection algorithm combined with Periodic Spikes Retrieval to the synthetic signals in Table \ref{table:signal}.
    If AutoPeriod detects a period on the residue signal of Periodic Spikes Retrieval, then Periodic Spikes Retrieval is applied again to this residue.
    We consider a detected period $P_d$ is correct if it is in a 5\% margin of the true period $P_t$, meaning $0.95P_t \leq P_d \leq 1.05P_t$.
    Table \ref{table:signal_results} displays the results.
    On the first application of AutoPeriod, the detected periods are correct for all cases.
    However, it does not detect the 2 periods for the cases G and H.
    Once the Spikes Retrieval has been applied, the AutoPeriod results are correct for all cases since it detects the remaining signal period (cases E to H), and it does not detect periodicity when there is none (cases A to D).

    \begin{table}
        \centering
        \resizebox{0.45\textwidth}{!}{%
            \begin{tabular}{|c|c|c|c|c|}
                \hline
                Case Name & RealPeriods & 1st AutoPeriod & 2nd AutoPeriod \\
                \hline
                A         & 48          & 48             & None           \\
                \hline
                B         & 48          & 48             & None           \\
                \hline
                C         & 23.6        & 24             & None           \\
                \hline
                D         & 23.6        & 24             & None           \\
                \hline
                E         & 48, 48      & 48             & 48             \\
                \hline
                F         & 48, 48      & 48             & 48             \\
                \hline
                G         & 21, 17      & 21             & 17             \\
                \hline
                H         & 23.6, 29.4  & 24             & 30             \\
                \hline
            \end{tabular}%
        }
        \caption{\centering AutoPeriod Results on the synthetic cases. The 2nd AutoPeriod column corresponds to AutoPeriod applied to the residue after Spikes Retrieval algorithm.}
        \label{table:signal_results}
    \end{table}

    In order to evaluate Spikes Retrieval results, the extracted signal is compared to the corresponding synthetic component using normalized mean absolute error.
    Results are presented in Table \ref{table:retrieval_results}.
    The algorithm excels in retrieving the most straightforward cases (A, B) even when there are two signals (E, F).
    When there is an offset in the period (C, D), the points created by this offset "disturbs" the spike average, leading to higher error.
    The algorithm has more trouble retrieving signals when their periods lead to overlaps (G, H).
    
    \begin{table}
        \centering
        \resizebox{0.33\textwidth}{!}{%
            \begin{tabular}{|c|c|c|c|c|}
                \hline
                Case Name & 1st nMAE               & 2nd nMAE      \\
                \hline
                A         & \textbf{0.04}          & None          \\
                \hline
                B         & \textbf{0.03}          & None          \\
                \hline
                C         & 0.19                   & None          \\
                \hline
                D         & 0.24                   & None          \\
                \hline
                E         & \textbf{0.03}          & \textbf{0.02}  \\
                \hline
                F         & \textbf{0.03}          & \textbf{0.06}  \\
                \hline
                G         & 0.32                   & 0.54            \\
                \hline
                H         & 0.28                   & 0.54            \\
                \hline
            \end{tabular}%
        }
        \caption{\centering Spikes Retrieval Results on the synthetic cases. The 2nd nMAE column corresponds to Spikes Retrieval applied to the residue of the first iteration.}
        \label{table:retrieval_results}
    \end{table}

    \begin{figure}
        \centering
        \includegraphics[scale=0.225]{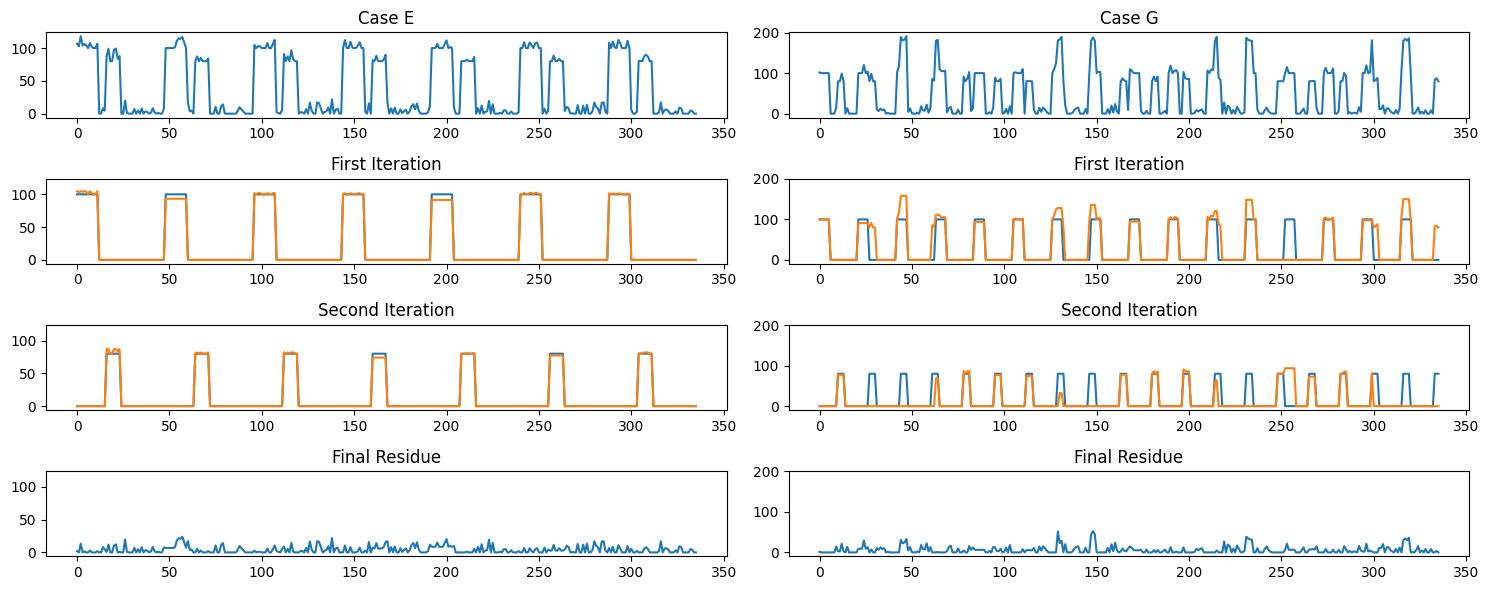}
        \caption{\centering Spikes Retrieval algorithm applied to cases E (left) and G (right). First plot is the input signal, last plot is the final residue. Plots 2 and 3 are the first and second iteration, found signal in blue and true component in orange.}
        \label{fig:results_cases}
    \end{figure}

    \section{Discussion}
    
    The methodology proposed in this article focuses on two tasks: holiday detection in low-frequency electricity consumption data, and the subsequent extraction of specific consumption patterns during holiday periods.
    This section examines the effectiveness of the proposed approach, its challenges, and its limitations.

    \paragraph{Effectiveness of the Methodology}
    \begin{itemize}
        \item Change Point Detection: The Bottom-Up algorithm demonstrated the ability to split low-frequency electricity consumption signals into appropriate segments.
        The selection of a proper threshold is vital, and the results were promising when it was fitted with correct criteria.
        \item Holiday Classification: The holiday classifiers outperformed the random classifier by a significant margin, with the variance-based classifiers (e.g., $N_{var}$ and $F_{var}$) exhibiting better performance. This indicates that variance is a more discriminative feature in identifying holidays.
        \item Periodic Spikes Retrieval: The AutoPeriod algorithm effectively detected periods in synthetic signals, and its combination with the Spikes Retrieval algorithm provided correct results for a variety of cases. In most instances, it identified the true period, but challenges arose when signals overlapped.
    \end{itemize}

    \paragraph{Challenges and Limitations}
    \begin{itemize}
        \item Data Labeling: In this article, we did not have access to precise labeling of holiday periods. We decided to label the sub-segments that were the most relevant, but a larger dataset with labels is necessary to evaluate our methods more precisely. Large-scale surveys sent to Hello Watt users could be a nice way to create a labeled dataset.
        \item Generalization to Real Data: While the AutoPeriod and Periodic Spikes Retrieval methods showed promising results on synthetic data and a subset of real-world data, further evaluation on a larger, diverse dataset is needed to confirm is applicability in various scenarios.
        \item Hyperparameter Sensitivity: Our methods (Bottom-Up algorithm and Holiday classifiers) heavily rely on hyperparameters. Training on a large dataset could be needed to generalize their use.
        \item Overlapping Signals: Periodic Spikes retrieval has great performance on sums of separated signals, but has trouble when there are overlaps.
    \end{itemize}

    \section{Conclusion}
    In conclusion, this research presents a novel methodology for detecting holidays and extracting consumption patterns in low-frequency electricity consumption data.
    The discussed algorithms, including Bottom-Up, holiday classifiers, AutoPeriod, and Periodic Spikes Retrieval, offer practical insights for optimizing energy use, and filtering out holidays for training other models.
    The potential benefits span residential consumers and utility companies, with implications for energy cost reduction and sustainable practices.
    As we look to the future, there is a need for further research to enhance the methodology's applicability to diverse settings, by creating a large and varied dataset.
    This work contributes to the ongoing efforts to improve energy management and environmental sustainability by allowing users to have better knowledge on their electricity usage.

    \bibliographystyle{elsarticle-num}
    \bibliography{references}

\end{document}